%% file: DSAA.tex
\documentclass[conference]{IEEEtran}
\IEEEoverridecommandlockouts
\usepackage{cite}
\usepackage{amsmath,amssymb,amsfonts}
\usepackage{algorithmic}
\usepackage{graphicx}
\usepackage{textcomp}
\usepackage{xcolor}
\usepackage{color}
\usepackage{tabularray}
\usepackage{subcaption}
\usepackage{float}
\usepackage{url}

\def\BibTeX{{\rm B\kern-.05em{\sc i\kern-.025em b}\kern-.08em
    T\kern-.1667em\lower.7ex\hbox{E}\kern-.125emX}}
\begin{document}

\title{MIS-ME: A Multi-modal Framework for \\ Soil Moisture Estimation*
\thanks{*This work is funded by USDA: Agriculture and Food Research Initiative (AFRI) - Data Science for Food and Agriculture Systems (DSFAS); Award Number:2023-67022-40019}
}


\author{
\IEEEauthorblockN{Mohammed Rakib\IEEEauthorrefmark{1}, Adil Aman Mohammed\IEEEauthorrefmark{1}, Cole Diggins\IEEEauthorrefmark{2}, Sumit Sharma\IEEEauthorrefmark{2},\\ Jeff Michael Sadler\IEEEauthorrefmark{2}, Tyson Ochsner\IEEEauthorrefmark{2}, Arun Bagavathi\IEEEauthorrefmark{1}}
\IEEEauthorblockA{\IEEEauthorrefmark{1}Department of Computer Science, \IEEEauthorrefmark{2}Department of Plant and Soil Sciences\\ Oklahoma State University, Stillwater, Oklahoma, United States\\
\{mohammed.rakib, adil.mohammed, ddiggin, sumit.sharma, jeff.sadler, tyson.ochsner, abagava\}@okstate.edu}
}

\maketitle


\begin{abstract}
Soil moisture estimation is an important task to enable precision agriculture in creating optimal plans for irrigation, fertilization, and harvest. It is common to utilize statistical and machine learning models to estimate soil moisture from traditional data sources such as weather forecasts, soil properties, and crop properties. However, there is a growing interest in utilizing aerial and geospatial imagery to estimate soil moisture. Although these images capture high-resolution crop details, they are expensive to curate and challenging to interpret. 
\textit{Imagine}, an AI-enhanced software tool that predicts soil moisture using visual cues captured by smartphones and statistical data given by weather forecasts. This work is a first step towards that goal of developing a multi-modal approach for soil moisture estimation. 
In particular, we curate a dataset consisting of real-world images taken from ground stations and their corresponding weather data. We also propose \emph{MIS-ME} - Meteorological \& Image based Soil Moisture Estimator, a multi-modal framework for soil moisture estimation. Our extensive analysis shows that MIS-ME achieves a MAPE of 10.14\%, outperforming traditional unimodal approaches with a reduction of 3.25\% in MAPE for meteorological data and 2.15\% in MAPE for image data, highlighting the effectiveness of tailored multi-modal approaches. Our source code will be available at \url{https://github.com/OSU-Complex-Systems/MIS-ME.git}


\end{abstract}

\input{1-Introduction}
\input{2-RelatedWorks}

\input{3-Dataset}
\input{4-Methodology}

\input{5-Results}
\input{6-Conclusion}

\bibliographystyle{IEEEtran}
\bibliography{IEEEabrv, DSAA}

\end{document}

%% file: 1-Introduction.tex
\section{Introduction}
Soil moisture is a key indicator of water usage and availability in agricultural fields\cite{Ochsner2013}. Soil moisture levels determine the amount of irrigation that is needed in irrigated fields, and it informs yield prediction in rain-fed fields. When there is insufficient soil moisture, crop productivity suffers, causing a significant financial burden for farmers. Although crucial, accurately measuring soil moisture requires in-situ sensor installations, which are difficult to scale and introduce upfront and ongoing maintenance costs \cite{PHILLIPS201473, wuacm, fengacm}. These difficulties are accentuated by ongoing efforts to develop and adopt precision agriculture. Precision agriculture, which uses technology to assist with sustainable agriculture management, though not a new idea \cite{PIERCE19991}, has gained increased traction in the past decade with advances in data availability, hardware (internet-connected devices), and data algorithms \cite{Bhat2021}. AI-based data-driven methods are one of the precision agriculture techniques that has proven useful in multiple applications like soil management, optimal crop-producing conditions, and determining watering quantities and fertilizers~\cite{sharma2020machine}. Developing such data-driven models for accurate soil moisture predictions, therefore, could provide useful information to farmers regarding locations that are unmonitored. In this work, we focus on utilizing machine learning approaches with digital images and weather observations collected from ground stations to estimate soil moisture to advance AI-driven precision agriculture.

\begin{figure}[tbp]
\centering
\centerline{\includegraphics[scale=0.035]{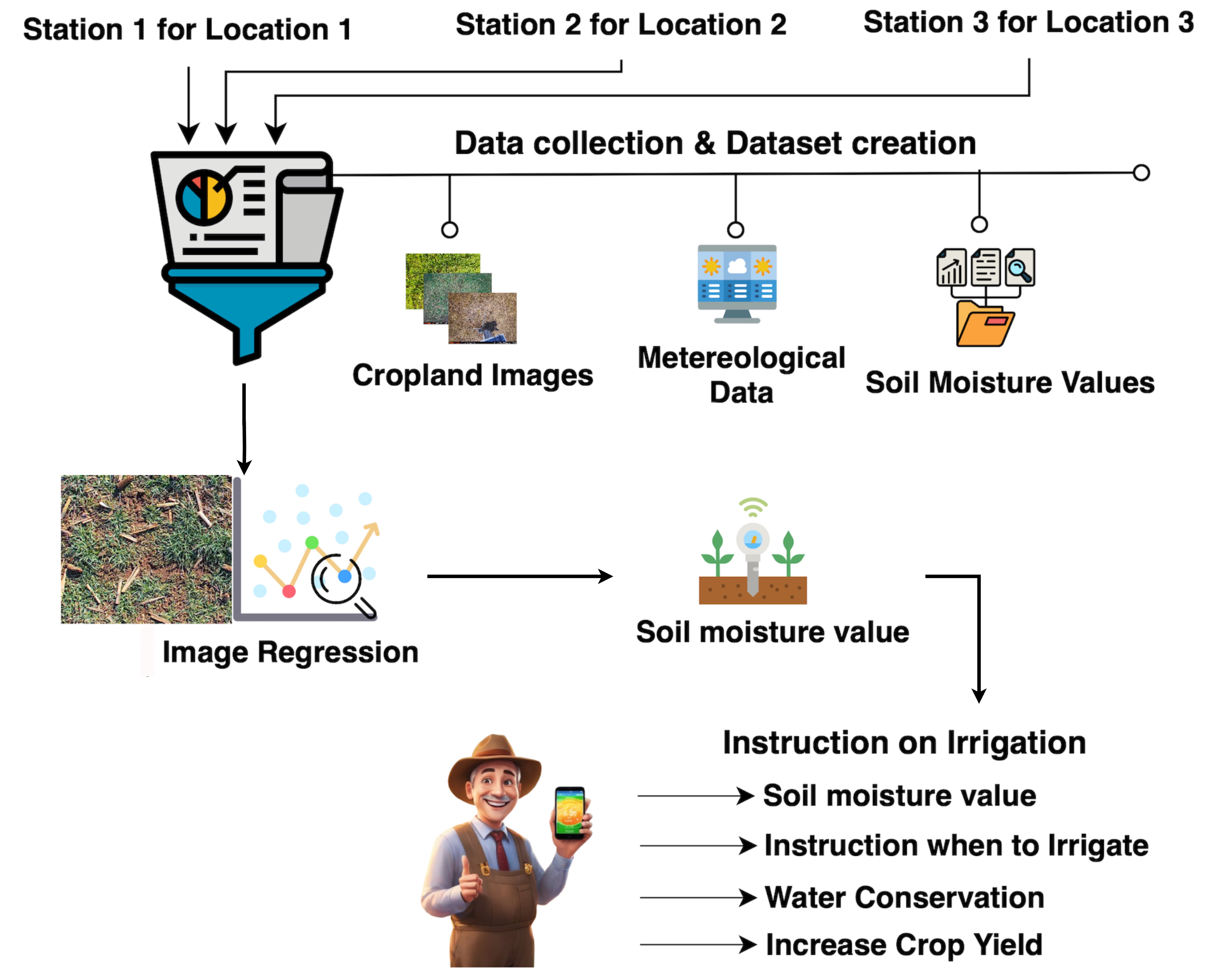}}
\caption{Integrating cropland images and meteorological data for improved soil moisture estimation and irrigation guidance.}
\label{fig:Flow chart}
\end{figure}

Soil moisture prediction has become a ubiquitous sub-task in the computational agriculture domain with the advancements in machine learning and deep learning algorithms~\cite{filipovic2022regional,shokati2023}. Multiple applied research directions for utilizing machine learning approaches in the soil moisture prediction sub-task have been explored in the literature. The most common approach is to utilize meteorological data,  weather forecasts, geographical details, and soil properties to predict the soil moisture level ahead in the future~\cite{yu2021hybrid,orth2021global,filipovic2022regional}. These methods utilize traditional machine learning models like linear regression, neural network regressor, and random forest models for such predictions. The second spectrum of research focuses on soil moisture prediction on either satellite imagery~\cite{singh2023deep} or UAV-based multispectral and thermal images~\cite{bertalan2022uav}. Such images offer a high-level understanding of the farmland and capture macro details of the area. However, these datasets are costly to procure, often capture only information about soil, and do not characterize micro details about crops and their relation to the soil moisture. Some of the works that study such micro details in soil moisture prediction sub-tasks consider images collected from a lab setting, which are less noisy \cite{mansur2022image, kim2023convolutional,sagayaraj2021determination}. Estimating soil moisture using photos taken from mobile phones or field cameras could be an impactful development towards real-time precision agriculture. As depicted in Fig.~\ref{fig:Flow chart}, we present an initial work towards precision agriculture with multi-modal data and develop methods to understand patterns in soil moisture to assist farmers in real time. Particularly, we contribute to analyzing the capacities of machine learning models on soil moisture estimation tasks directly on images taken from field cameras along with meteorological weather observations. We study soil moisture in terms of volumetric water content (\textit{VWC}), which is the proportion of water volume to soil volume, usually expressed as cm\textsuperscript{3}cm\textsuperscript{-3} \cite{task2016evaporation}. 



We present \textbf{\textit{three-fold}} contributions in this work:

\begin{itemize}
\item We curate soil patch images from the existing photos collected from ground stations along with associated meteorological data, with an aim to enhance the performance of soil moisture or \textit{VWC} estimation problems.
\item We evaluate the potential of image regression models to predict the \textit{VWC} from our pre-processed datasets. To our knowledge, this is the first effort to quantify soil moisture directly from raw cropland images.
\item We introduce the \emph{MIS-ME} framework, featuring three innovative multi-modal approaches for soil moisture estimation: MIS-ME with Multi-modal Concat, MIS-ME with Hybrid Loss, and MIS-ME with Learnable Parameters. Each approach is designed to leverage both image features from soil patches and corresponding meteorological data, optimizing prediction accuracy through distinct combination and weighting strategies. The code will be made available after the review process.
\end{itemize}

%% file: 2-RelatedWorks.tex
\section{Related Work}
\subsection{Meteorological Data in Soil Moisture Prediction}
Current trends in soil moisture forecasting display a broad spectrum of applications for meteorological data, ranging from statistical models to deep learning methodologies. The Seasonal ARIMA model, when coupled with a water balance equation, shows significant accuracy in predicting soil moisture across various depths \cite{fu2023}, proving effective especially at depths minimally affected by external variables. Moving to machine learning, \cite{prakash2018} employs multiple linear regression and support vector regression, showcasing the effectiveness of linear regression in specific contexts. LightGBM's potential in high-resolution soil moisture prediction outperforms traditional methods like linear regression and random forest, especially in IoT-enabled soil moisture data scenarios~\cite{togneri2022}.

Deep learning has been widely used for soil moisture prediction with models like LSTMs, 1D-CNNs, encoders, and even the fusion of CNNs with LSTMS~\cite{cnn-with-lstm}.  These models effectively handle diverse soil moisture data under varying conditions. \cite{cai2019} proposes a Deep Neural Network Regression (DNNR) model, noted for its robust data fitting and generalization capabilities in soil moisture prediction. Additionally, \cite{li2022improving} introduces the EDT-LSTM model, enhancing prediction accuracy by focusing on intermediate time-series data. Similarly, \cite{yu2021hybrid} presents a hybrid CNN-GRU model tailored for maize root zone moisture prediction, which excels in accuracy and convergence rate. Collectively, these studies underscore the rapidly evolving landscape of soil moisture prediction methodologies, each contributing unique insights and advancements to the field.

\subsection{Image-Based Soil Moisture Estimation}
The application of satellite and UAV imagery in soil moisture prediction has revolutionized the field, utilizing various methodologies to improve precision and utility. The first approach involves extracting features from satellite images for soil moisture prediction. Studies like \cite{celik2022soil} and \cite{singh2023deep} demonstrate this technique by integrating features from Sentinel-1 backscatter data, soil moisture active passive data, and topographic information within deep learning frameworks. Similarly, \cite{hegazi2023prediction} focuses on the effective use of Sentinel-2 bands and indices like NDWI and NDVI, utilizing a convolutional neural network for accurate soil moisture estimation. Additionally, \cite{habiboullah2023} introduces models based on ConvLSTM layers and visual transformers, employing NDVI and NSMI satellite data for soil moisture content prediction, highlighting the effectiveness of deep learning in processing satellite-derived information. In the second approach, UAV imagery is employed for soil moisture prediction, offering high-resolution data essential for precision agriculture \cite{shokati2023, dingacm}. \cite{shokati2023} explores the use of visible UAV imagery combined with models like Random Forest and Multilayer Perceptron, demonstrating the potential of aerial imagery in bare soil field assessments. Lastly, lab-based image analysis offers a controlled environment for soil moisture prediction. \cite{mansur2022image}, \cite{kim2023convolutional}, and \cite{sagayaraj2021determination} utilize soil surface images within laboratory settings to develop predictive models. These studies apply image processing techniques and convolutional neural networks to analyze soil water content and density, showcasing the capabilities of image processing in soil analysis under controlled conditions.

\subsection{Our Contribution}
In contrast to existing research primarily relying on controlled lab settings for soil image acquisition, our analysis uses raw images captured in natural environments such as field node cameras. Our methodology stands out by integrating these real-world soil images with corresponding meteorological data to enhance the prediction of soil moisture. Moreover, we introduce the three-way \emph{MIS-ME} framework, which presents three fusion techniques to combine soil patch images and tabular meteorological data. Our method of utilizing raw soil images captured in the wild positions our research as a distinctive and impactful contribution towards precision agriculture.


%% file: 3-Dataset.tex
\section{Dataset}
\label{sect:dataset}

In this work, we use field camera images and associated meteorological data for machine-learning tasks. The dataset was collected as a part of ongoing research to develop improved cropland monitoring stations. These stations are equipped with cosmic-ray neutron detectors for non-contact, field-scale soil moisture monitoring, downward-facing outdoor cameras for visually monitoring soil and crop conditions, and all-in-one weather stations for recording meteorological data. 
The stations were deployed in Chickasha (\emph{Station1}), Chattanooga (\emph{Station2}), and Braman (\emph{Station3}) in Oklahoma from 2020 to 2021 to collect atmospheric data and cropland images.
The stations were deployed in Chickasha (\emph{Station1}), Chattanooga (\emph{Station2}), and Braman (\emph{Station3}) in Oklahoma from 2020 to 2021 to collect atmospheric data and cropland images. The meteorological data include air temperature, relative humidity, and rainfall. Images were collected at regular intervals from 9 am to 5 pm. These data capture the dynamic nature of soil moisture, reflecting the complexity and variability of agricultural croplands. 

\begin{table}[tbp]
\centering
\caption{Statistical overview of the soil moisture dataset}
\label{tab:datasetinfo}
\resizebox{\linewidth}{!}{%
\begin{tblr}{
  cells = {c},
  hlines,
  vlines,
  hline{3-4} = {-}{dashed},
}
{\textbf{Station }\\\textbf{Name}} & {\textbf{Sand-}\\\textbf{Silt-}\\\textbf{Clay}\\\textbf{(\%})} & {\textbf{Crop}\\\textbf{land}\\\textbf{Images}} & {\textbf{Soil }\\\textbf{Patches}} & {\textbf{Days }\\\textbf{Count}} & {\textbf{\textit{VWC}}\\\textbf{Range~}\\\textbf{\textbf{(cm\textsuperscript{3}cm\textsuperscript{-3})}}} & {\textbf{\textit{VWC}}\\\textbf{Mean~}\\\textbf{\textbf{(cm\textsuperscript{3}cm\textsuperscript{-3})}}} & {\textbf{\textit{VWC}}\\\textbf{STD~}\\\textbf{\textbf{(cm\textsuperscript{3}cm\textsuperscript{-3})}}} \\
Station1                           & {18.0 -\\56.5 -\\25.6}                                         & 977                                             & 1822                               & 202                              & {0.158 -\\0.417}                                                                                 & 0.3085                                                                                          & 0.0496                                                                                         \\
Station2                           & {28.8 -\\45.2 -\\26.0}                                         & 704                                             & 1604                               & 246                              & {0.118 -\\0.435}                                                                                 & 0.2455                                                                                          & 0.0646                                                                                         \\
Station3                           & {19.5 -\\54.2 -\\26.3}                                         & 925                                             & 3356                               & 258                              & {0.151 -\\0.423}                                                                                 & 0.3126                                                                                          & 0.0582                                                                                         
\end{tblr}
}
\end{table}

Table~\ref{tab:datasetinfo} presents a statistical overview of the raw dataset, encompassing data from three stations used in this work. The table specifies the respective soil compositions in terms of sand, silt, and clay. Notably, all stations exhibit similar clay content, with marginal variations in sand and silt ratios, particularly at \emph{Station2}. Additionally, the table illustrates the range, mean, and standard deviation of volumetric water content (\textit{VWC}) at each station. The \textit{VWC} across the stations varies from approximately 0.1 to 0.4 cm\textsuperscript{3}cm\textsuperscript{-3}. Both \emph{Station1} and \emph{Station3} show similar mean values and standard deviations for soil moisture, whereas \emph{Station2} exhibits a marginally lower mean and the highest standard deviation. This indicates that while \emph{Station1} and \emph{Station3} share comparable soil characteristics, \emph{Station2} differs slightly in its soil properties. 


\begin{figure}[bp]
   \centering
   \includegraphics[width=\linewidth]{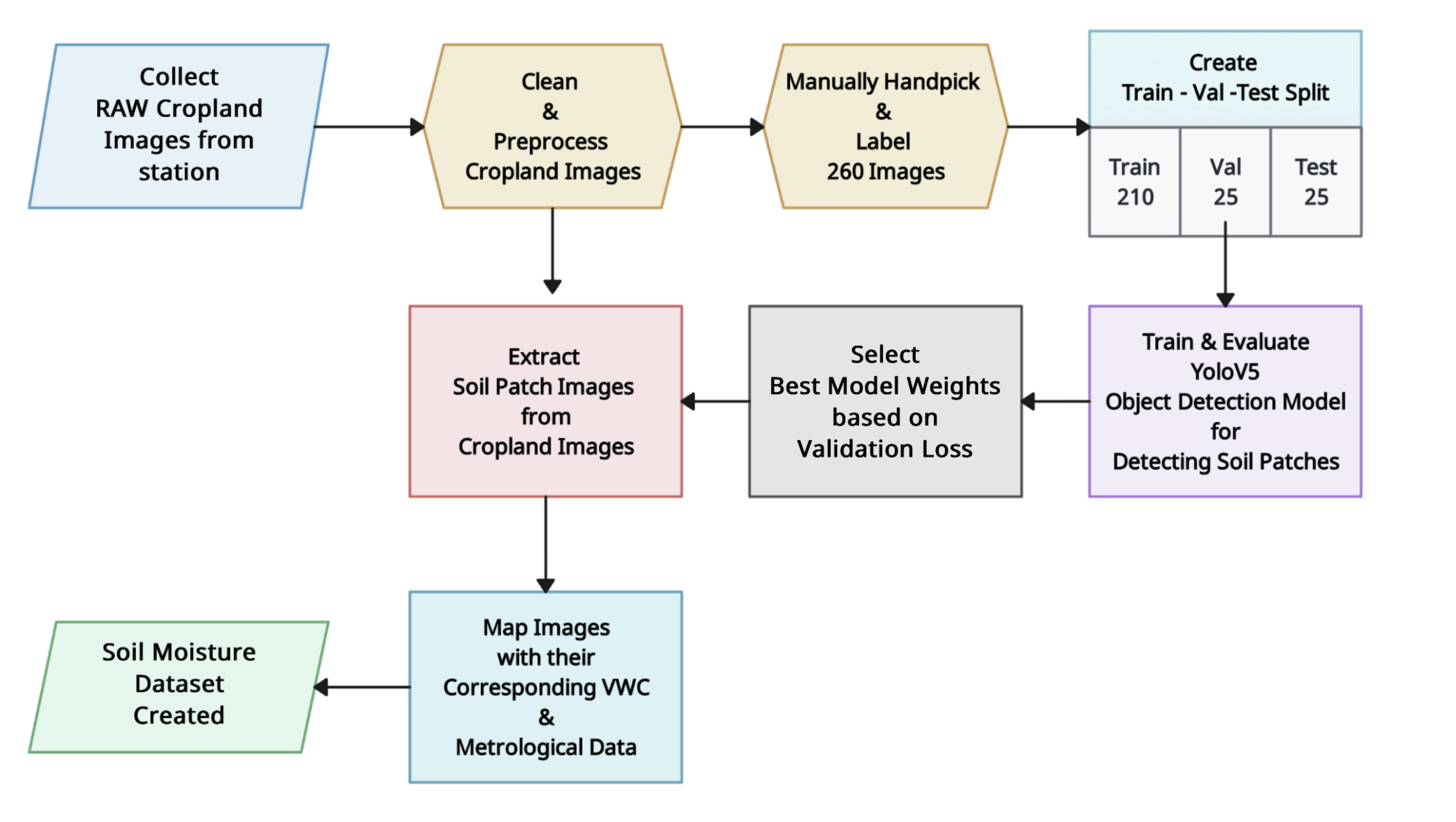}
   \caption{Overview of the dataset creation pipeline}
   \label{fig:datasetflow}
\end{figure}

\subsection{Preprocessing Cropland Images for Soil Patch Extraction}
Images available in our dataset are collected from crop fields, unlike existing images-based soil moisture prediction methods \cite{mansur2022image, kim2023convolutional, sagayaraj2021determination}, which collect soil samples from a lab setting. Hence, our raw cropland images include not only soil patches but also objects like crops, shadows, blurred images, and night images. Initially, we had 1950 images for \emph{Station1}, 1404 images for \emph{Station2}, and 1835 images for \emph{Station3}. After manually removing noisy images, we have a total of 2606 raw images with 977 images from \emph{Station1}, 704 images from \emph{Station2}, and 925 images from \emph{Station3}. 
In this work, we follow a 3-step approach to prepare these filtered cropland images for machine learning tasks: (i) handpick and label (draw bounding box over soil patches) 250 cropland images, (ii) train \& evaluate YOLOv5 \cite{yolov5} using those images and (iii) extract soil patches using the trained YOLOv5 from the filtered images of the three stations. Fig.~\ref{fig:datasetflow} gives a complete overview of the dataset creation process.

\begin{table}
\centering
\caption{Evaluation metrics for YOLOv5 model}
\label{tab:yoloeval}
\resizebox{\linewidth}{!}{%
\begin{tblr}{
  cells = {c},
  cell{2}{1} = {r=3}{},
  cell{5}{1} = {r=3}{},
  cell{8}{1} = {r=3}{},
  vlines,
  hline{1-2,11} = {-}{},
  hline{3-10} = {2-6}{},
}
\textbf{Data Split} & \textbf{Evaluation criteria} & \textbf{Precision} & \textbf{Recall} & \textbf{F1 Score} & \textbf{AP} \\
Train               & IoU@0.5                      & 0.952              & 0.822           & 0.882             & 0.881       \\
                    & IoU@0.75                     & 0.955              & 0.919           & 0.936             & 0.969       \\
                    & IoU@0.90                     & 0.851              & 0.814           & 0.832             & 0.896       \\
Validation          & IoU@0.5                      & 0.830              & 0.706           & 0.763             & 0.710       \\
                    & IoU@0.75                     & 0.830              & 0.706           & 0.763           & 0.704       \\
                    & IoU@0.90                     & 0.801              & 0.688           & 0.740             & 0.680       \\
Test                & IoU@0.5                      & 0.818              & 0.685           & 0.745             & 0.716       \\
                    & IoU@0.75                     & 0.805              & 0.682           & 0.738             & 0.693       \\
                    & IoU@0.90                     & 0.739              & 0.659           & 0.697             & 0.651       
\end{tblr}
}
\end{table}

First, we manually pick \emph{260} raw images and draw bounding boxes to mark soil patches. This small sample is split into three categories: train (210 images), validation (25 images), and test (25 Images) sets.
Next, we train and evaluate the YOLOv5 object detection model to extract soil patches from all preprocessed images. Table~\ref{tab:yoloeval} shows the performance of the finetuned YOLOv5 model with Intersection over Union (IoU) evaluation criteria thresholds at 0.5, 0.75, and 0.90. 
Observing Table~\ref{tab:yoloeval}, we see that the model performs well across all overlapping criteria for the training, validation, and test sets. The precision, recall, F1 score, and AP remain steady and show similar trends across all the dataset splits. The evaluation results indicate that the trained YOLOv5 model is effective in detecting soil patches with consistent performance across different IoU thresholds and data splits. Subsequently, we use this trained \emph{YOLOv5} model to extract a total of 6782 soil patches from all our preprocessed 2606 raw images across the three stations and create the dataset. 
We selected a confidence threshold of 0.5 for the YOLOv5 model to strike a balance between minimizing false positives and maximizing the detection of soil patches.
After extracting soil patches and eliminating irrelevant images, \emph{Station3} has the highest soil patch quantity with 3356 patches, followed by 1822 patches for \emph{Station1} and 1604 for \emph{Station2}. 
\begin{table}[tbp]
\centering
\caption{Description of Meteorological Variables}
\label{tab:datasetsymbol}
\resizebox{\linewidth}{!}{%
\begin{tblr}{
  cells = {c},
  vlines,
  hline{1-2,18} = {-}{},
}
\textbf{Abbreviation}   & \textbf{Variable Description}                 & \textbf{Unit} \\
$ T_\text{air} $       & air temperature                            & \textdegree{}C \\
$ T_\text{mod} $       & temperature of the internal system module     & \textdegree{}C \\
$ T_\text{hs} $        & temperature of the humidity sensor            & \textdegree{}C \\
$ RH $                 & percent relative humidity                     & \%             \\
$ RH_\text{mod} $      & humidity of the internal system module        & \%             \\
$ P $                  & precipitation                            & mm             \\
$ \Phi_\text{solar} $  & solar radiation flux given by ClimaVUE 50 sensor & $W/m^2$          \\
$ P_\text{vapor} $     & pressure exerted by water vapor in air        & kPa            \\
$ P_\text{bar} $       & barometric pressure                           & hPa            \\
$ v_\text{wind} $      & wind speed                             & m/s            \\
$ v_\text{gust} $      & highest speed of wind during short bursts     & m/s            \\
$ v_\text{north} $     & north relative wind speed                     & m/s            \\
$ v_\text{east} $      & east relative wind speed                      & m/s            \\
$ \theta_\text{wind} $ & direction of the wind                         & degrees        \\
$ Tilt_\text{NS} $     & north south tilt of the station               & degrees        \\
$ Tilt_\text{WE} $     & east west tilt of the station                 & degrees        
\end{tblr}
}
\end{table}

\subsection{Preprocessing Meteorological Data}

\begin{figure}[bp]
\centerline{\includegraphics[scale=0.42]{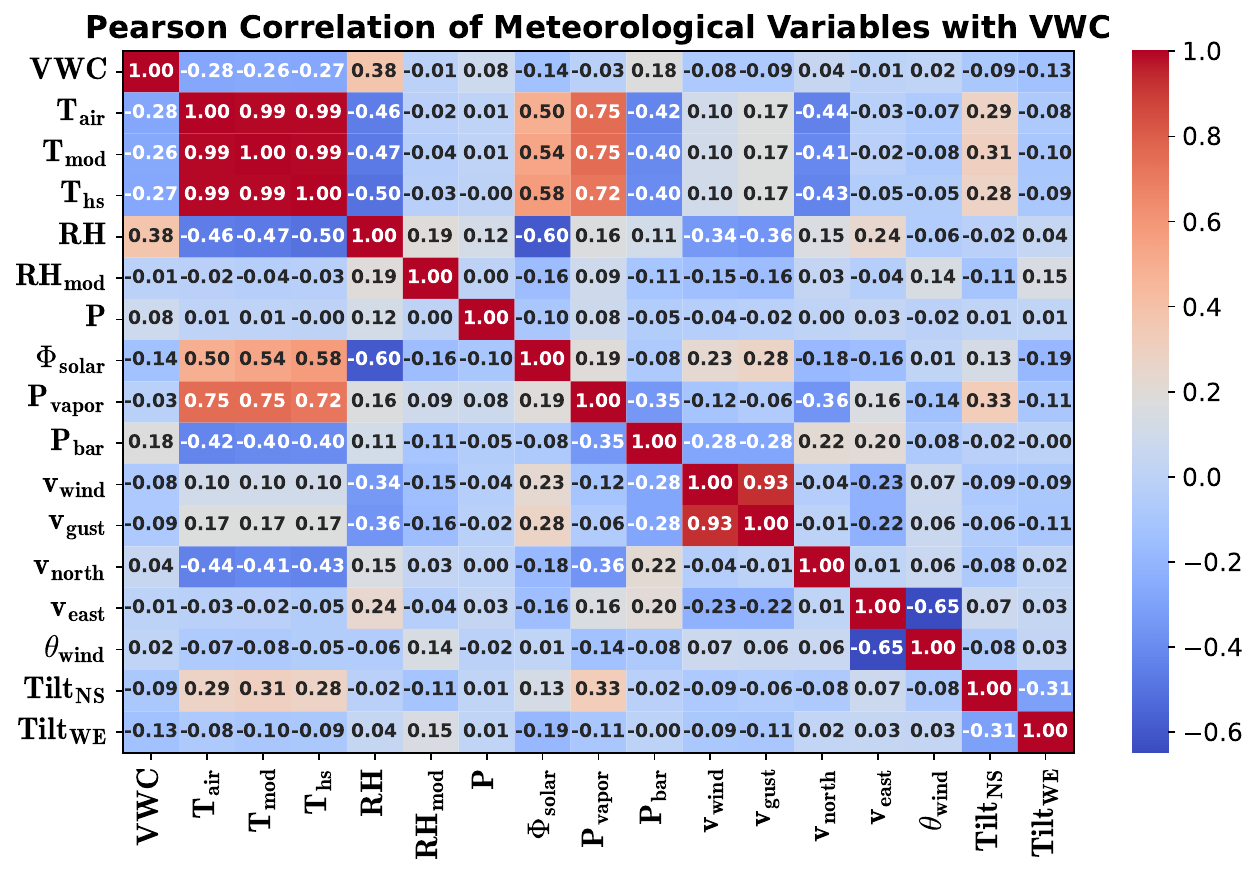}}
\caption{Pearson correlation heatmap of the 16 meteorological variables with \textit{VWC}.}
\label{fig:heatmap}
\end{figure}

Table~\ref{tab:datasetsymbol} provides an overview of the meteorological variables measured by the cropland monitoring stations with their respective units. While each variable offers valuable insights for meteorological studies, our research will specifically focus on variables that strongly correlate with \textit{VWC}. Fig.~\ref{fig:heatmap} displays the Pearson correlation analysis between \textit{VWC} and various meteorological variables. Notably, \textit{VWC} exhibits significant inverse correlations with \( T_{air} \), \( T_{mod} \), and \( T_{hs} \), with coefficients of -0.28, -0.26, and -0.27, respectively. In contrast, \textit{VWC} has a strong positive correlation of 0.38 with \( RH \). Due to the high intercorrelation among \( T_{air} \), \( T_{mod} \), and \( T_{hs} \), we retain only \( T_{air} \) and exclude the others in our dataset to enhance machine learning efficiency and reduce redundancy. \textit{VWC} also correlates linearly with \( P_{bar} \), \( \Phi_{solar} \), \( Tilt_{NS} \), \( Tilt_{WE} \), \( P \), \( v_{wind} \), and \( v_{gust} \), with respective coefficients of 0.18, 0.14, 0.09, 0.13, 0.08, 0.08, and 0.09. Due to the high correlation between \( v_{wind} \) and \( v_{gust} \), we choose to include only \( v_{wind} \). We exclude other variables, such as \( RH_{mod} \), \( P_{vapor} \), \( v_{north} \), \( v_{east} \), and \( \theta_{wind} \) due to minimal correlation with \textit{VWC}.

The selected meteorological features for our model training are \( T_{air} \), \( RH \), \( P \), \( P_{bar} \), \( \Phi_{solar} \), \( Tilt_{NS} \), \( Tilt_{WE} \), and \( v_{wind} \), focusing on the variables that most significantly impact \textit{VWC}. All the selected features are normalized using the standard z-score normalization technique. 

\begin{figure}[tbp]
\centerline{\includegraphics[scale=0.4]{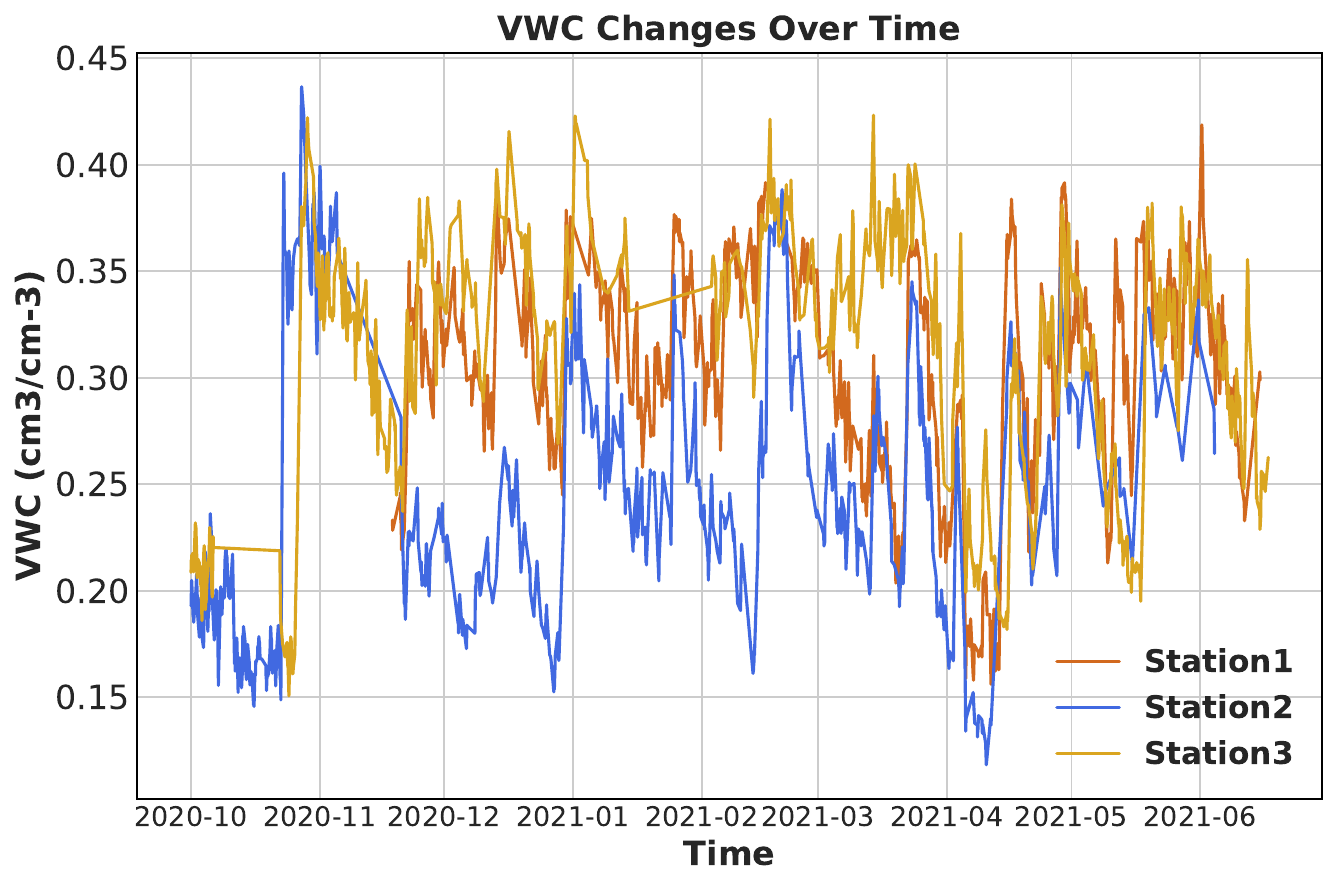}}
\caption{Trend of \textit{VWC} from October 2020 to June 2021 of the three stations with \emph{Station1} and \emph{Station3} showing similar trends compared to \emph{Station3}.}
\label{fig:vwctimeplot}
\end{figure}

\subsection{Dataset Analysis}
In our analysis of the \textit{VWC} in Fig.~\ref{fig:vwctimeplot}, we note that \emph{Station1} and \emph{Station3}, with similar soil types, show aligned \textit{VWC} peaks, likely linked to similar reactions to moisture changes during events like irrigation or rainfall. On the other hand, \emph{Station2}, characterized by a higher sand content, displayed sharper \textit{VWC} peaks and rapid declines, highlighting the influence of its soil texture on moisture fluctuations. This pattern, especially evident in \emph{Station2}, emphasizes the role of soil texture in managing water retention and drainage.

%% file: 4-Methodology.tex
\section{Methodology}
In this section, we formulate and describe the traditional unimodal approaches and our proposed three-way MIS-ME framework for soil moisture estimation.

\subsection{Image-Only Regression Approach}
This approach utilizes state-of-the-art image feature extractors such as ResNet18 \cite{resnet18}, InceptionV3 \cite{inceptionv3}, MobileNetV2 \cite{mobilenetv2}, and EfficientNetV2 \cite{efficientnetv2}, each offering unique advantages for image regression tasks. The image feature extraction process can be represented by Eq.~\ref{eq:img_feat}.
\begin{equation}
    ImageFeatureExtractor(X_{img}): \mathcal{F}(X_{img}) \rightarrow \mathbb{R}^n
    \label{eq:img_feat}
\end{equation}
Here, \( X_{img} \in \mathbb{R}^{w \times h \times 3} \) denotes the input soil patch image, \( \mathcal{F} \) represents the feature extraction function and it can be any of the state-of-the-art feature extraction models. $\mathbb{R}^n$ is an $n-$ dimensional feature vector of $X_{img}$. The extracted features are then mapped to the predicted \textit{VWC} value through a dense output layer, as given in Eq.~\ref{eq:reg_linear}.
\begin{equation}
    ImageRegressionLayer(\mathbb{R}^n): f(\mathbb{R}^n) \rightarrow \mathbb{R}
    \label{eq:reg_linear}
\end{equation}
where, \( f \) denotes the linear regression function.

\subsection{Meteo-Only Regression Approach}
\label{sec:msme}
This approach focuses on tabular meteorological data only for \textit{VWC} prediction. We introduce our own architecture for this approach titled Meteorological Soil Moisture Estimator (MSME). MSME has a series of fully connected layers with dropout and batch normalization to process the meteorological data efficiently. It is used in the MIS-ME framework to train on meteorological data. The meteorological feature extraction process for MSME is given in Eq.~\ref{eq:meteo_feat}.
\begin{equation}
\begin{split}
MSME(X_{met}):
\mathcal{M}(X_{met}) \rightarrow \mathbb{R}^m   
\end{split}  
\label{eq:meteo_feat}
\end{equation}
Here, \( X_{met} \in \mathbb{R}^{k} \) is the input meteorological data vector, \( \mathcal{M} \) represents sequential neural network operations, and \( m \) is the number of extracted features. These features are then fed into a linear regression layer to predict \textit{VWC}, as given in Eq.~\ref{eq:meteo_reg}.
\begin{equation}
MeteoRegressionLayer(\mathbb{R}^m): g(\mathbb{R}^m) \rightarrow \mathbb{R}
\label{eq:meteo_reg}
\end{equation}
where, \( g \) is the linear regression function.

\subsection{Meteorological \& Image based Soil-Moisture Estimator (MIS-ME) Framework}
We propose the MIS-ME framework to combine both image and meteorological data for enhanced knowledge-encoded \textit{VWC} prediction. We introduce three novel multimodal approaches for our MIS-ME framework, that have not been explored before in the field of soil moisture estimation. The approaches are formalized below utilizing formulations given in Eq.~\ref{eq:img_feat} and Eq.~\ref{eq:meteo_feat}.

\begin{figure}[tbp]
\centering
\centerline{\includegraphics[scale=0.50]{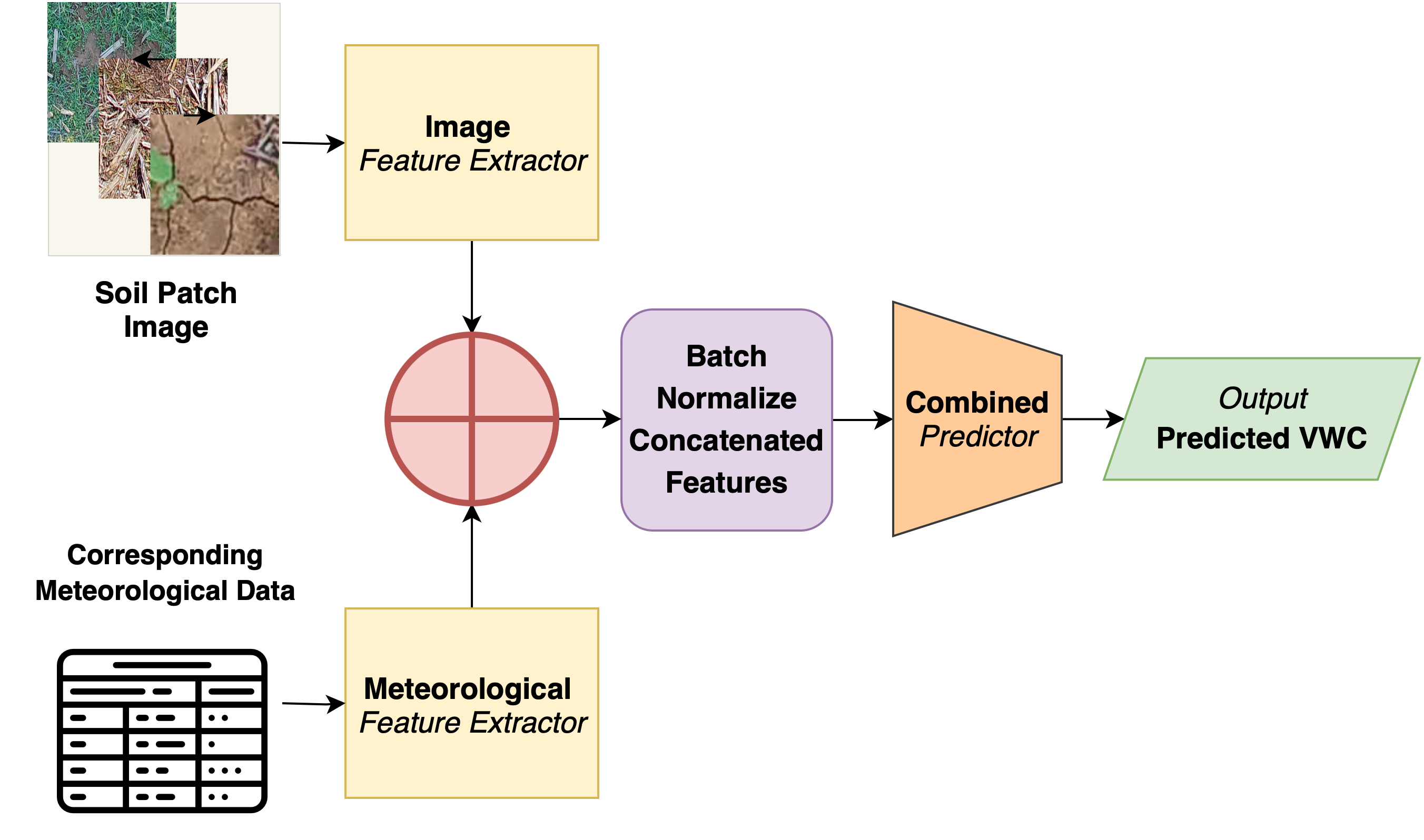}}
\caption{MIS-ME with Multimodal Concat extracts soil patch image features using a trained image feature extractor and extracts tabular meteorological features using the proposed MSME model in \ref{sec:msme}. The extracted features are combined and batch-normalized to pass through a series of fully connected layers for soil moisture regression.}
\label{fig:concatplot}
\end{figure}

\subsubsection{MIS-ME with Multimodal Concat}
The concatenation approach within the MIS-ME framework directly combines the feature vectors obtained from both soil patch images and meteorological data. This method leverages the diverse feature representations directly for regression. The concatenated vector is then normalized and fed into a sequence of fully connected layers to predict the \textit{VWC}. The mathematical representation of the concatenation process is shown in Eq.~\ref{eq:normal_concat}.
\begin{equation}
\begin{split}
MISME_{concat}(X_{img}, X_{met}): \\
\mathcal{P} ( \mathcal{F}(X_{img}) \oplus \mathcal{M}(X_{met}) ) \rightarrow \mathbb{R}^{m+n}
\end{split}
\label{eq:normal_concat}
\end{equation}
Here, \(X_{img} \in \mathbb{R}^{w \times h \times 3}\) and \(X_{met} \in \mathbb{R}^{k}\) represent the input image and meteorological data respectively. The feature vectors \(\mathcal{F}(X_{img})\) and \(\mathcal{M}(X_{met})\) are extracted using their respective feature extractors. The \(\oplus\) operator denotes the concatenation of these vectors, combining them into a single feature vector. Next, we apply batch normalization to align the scale of the features from the two different sources. This is done to stabilize and smoothen the learning process of the model. The normalized combined features are then passed to the prediction function \(\mathcal{P}\), which applies a series of dense layers to output the predicted \textit{VWC}.

Fig.~\ref{fig:concatplot} illustrates the data flow through the model, showing how inputs are transformed into a single output through the concatenation and subsequent processing steps. This setup allows the model to learn from both types of data simultaneously, potentially capturing interactions between image-derived features and meteorological factors that are indicative of soil moisture.

\subsubsection{MIS-ME with Hybrid Loss}

\begin{figure}[tbp]
\centering
\centerline{\includegraphics[width=\linewidth]{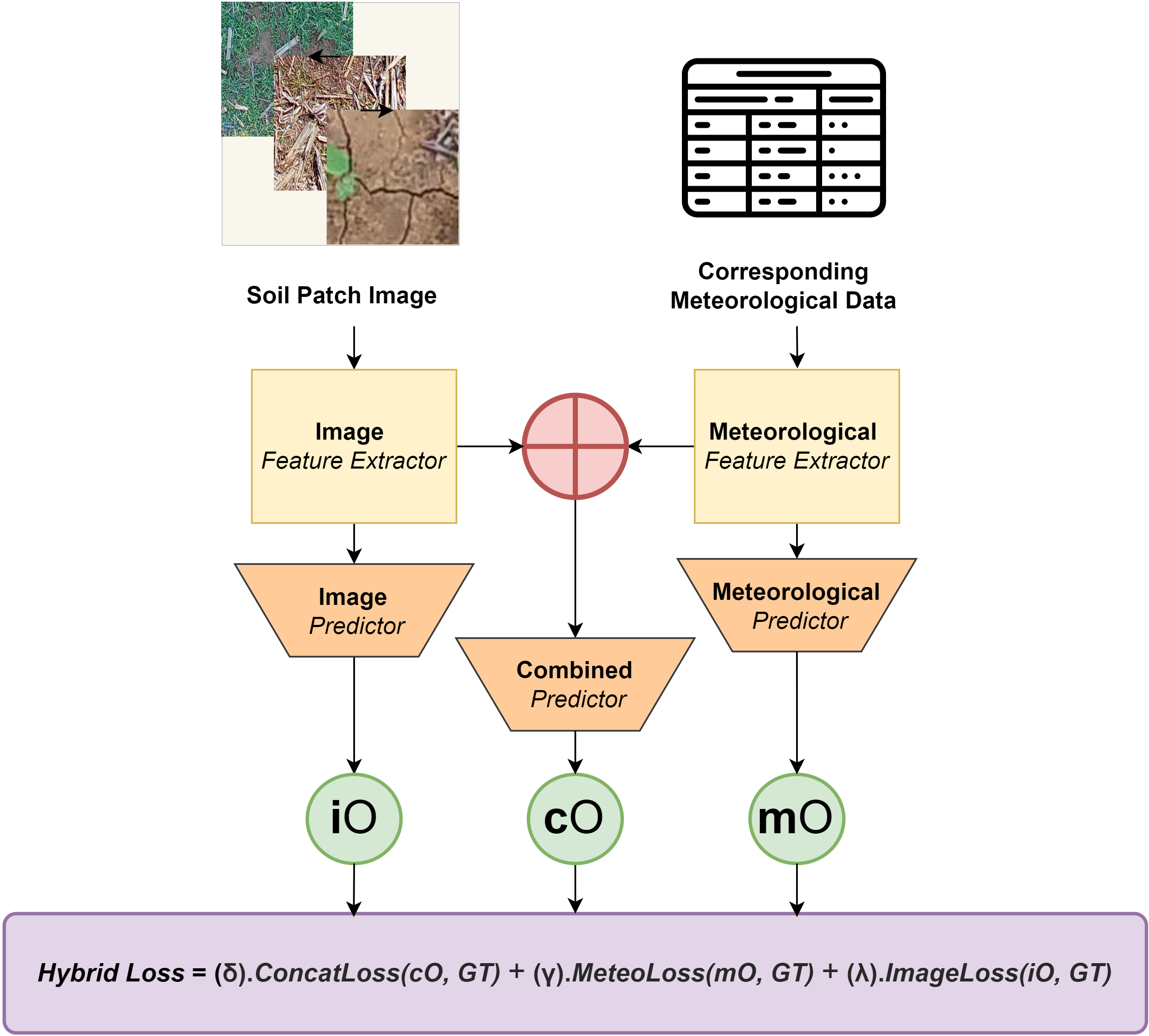}}
\caption{MIS-ME with Hybrid Loss differs from the Multimodal Concat approach by introducing two more predictors apart from the combined predictor and a hybrid loss function that combines multiple losses in a weighted manner.}
\label{fig:hybridplot}
\end{figure}

This approach employs a unique strategy by integrating three different loss functions designed to enhance the model's predictive accuracy and robustness across different types of input data. This multi-loss strategy harmonizes the losses from concatenated features, meteorological features, and image features, each weighted to optimize their contribution towards the final prediction. As illustrated in Fig.~\ref{fig:hybridplot}, this approach facilitates a holistic learning process that captures diverse characteristics of both soil patch images and meteorological data. The mathematical formulation of our triplet loss is given in Eq.~\ref{eq:hyrbid_loss}.

\begin{equation}
\begin{split}
    \mathcal{L}_{hybrid} = \delta \cdot \mathcal{L}_{concat}(cO, GT) + 
    \gamma \cdot \mathcal{L}_{meteo}(mO, GT) + \\ \lambda \cdot \mathcal{L}_{image}(iO, GT)
\end{split}   
\label{eq:hyrbid_loss}
\end{equation}
where:
\begin{itemize}
    \item \( \mathcal{L}_{concat} \) is the loss computed from the concatenated output (cO) against the ground truth (GT).
    \item \( \mathcal{L}_{meteo} \) is the loss from the meteorological output (mO).
    \item \( \mathcal{L}_{image} \) is the loss calculated from the image output(iO).
    \item \( \delta \), \( \gamma \), and \(\lambda\) are the weighting coefficients for the concatenated loss, meteo loss, and image loss, respectively.
\end{itemize}

To the best of our knowledge, this application of a triplet loss framework, incorporating a dynamic weighting of losses across multiple modalities, represents a novel approach in the field of multimodal soil moisture estimation. By leveraging this method, our model not only adapts to the inherent variability in the data sources but also significantly enhances the robustness of the predictions by systematically balancing the influence of different feature sets.

\subsubsection{MIS-ME with Learnable Parameter}
\begin{figure}[tbp]
\centering
\centerline{\includegraphics[scale=0.50]
{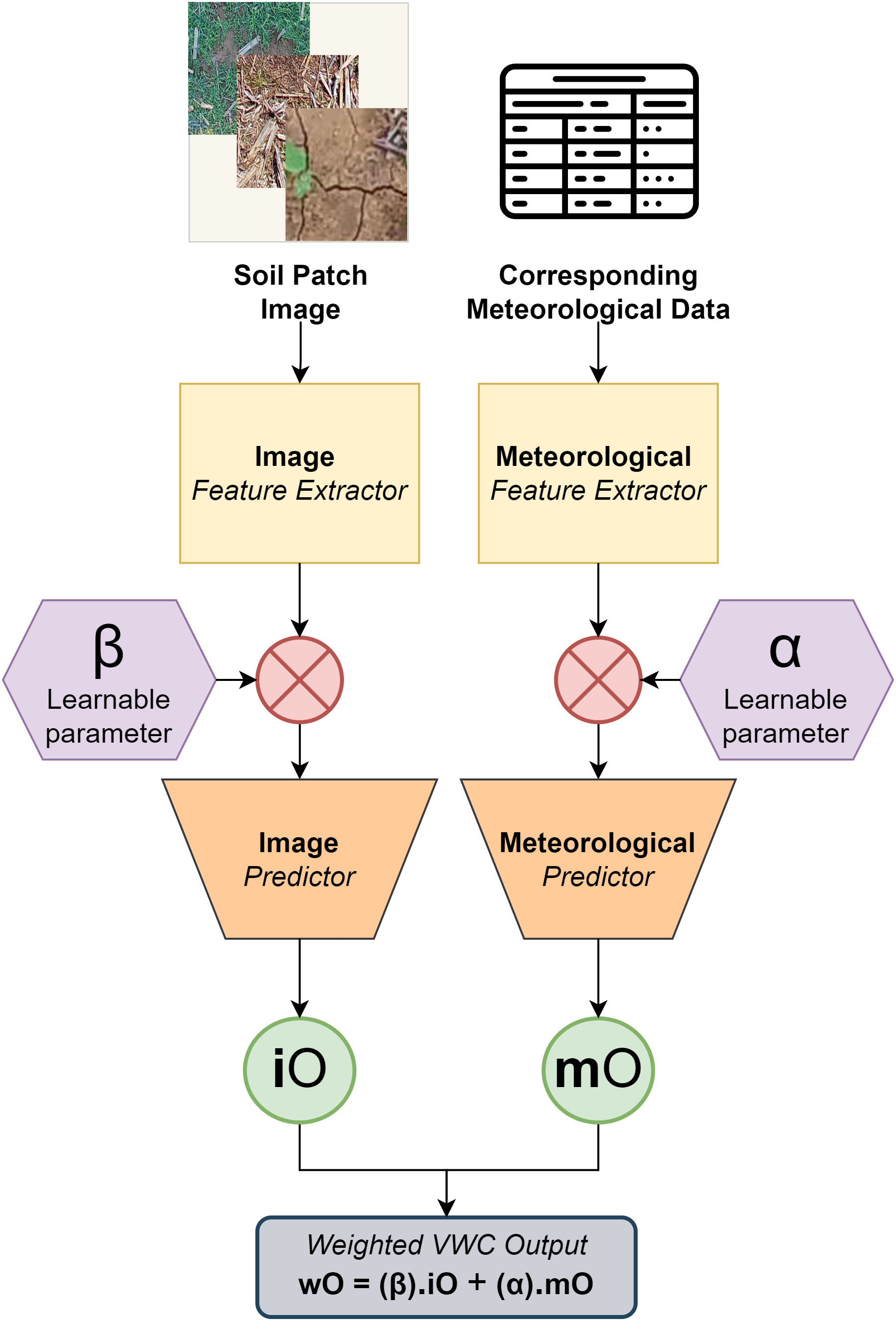}}
\caption{MIS-ME with Learnable Parameter introduces two learnable parameters: $\beta$ for images and $\alpha$ for meteo data, each of which gets multiplied with the output from their respective feature extractors. The final \textit{VWC} output is then weighted using the learned value of these two parameters.}
\label{fig:lpplot}
\end{figure}

This approach utilizes two distinct learnable coefficients, \( \alpha \) and \( \beta \), to dynamically adjust the influence of meteorological and image-derived features on the final prediction, as shown in Fig.~\ref{fig:lpplot}. This method enables the model to adaptively allocate more weight to the modality that provides more predictive power, optimizing the integration of different data sources for soil moisture estimation. The formulation for the model prediction using two learnable parameters is expressed in Eq.~\ref{eq:two_lp}.

\begin{equation}
    \hat{y} = \alpha \cdot \mathcal{P}_{meteo}(mO) + \beta \cdot \mathcal{P}_{image}(iO)
\label{eq:two_lp}
\end{equation}
where:
\begin{itemize}
    \item \( \mathcal{P}_{meteo}(mO) \) represents the prediction derived using meteorological data
    \item \( \mathcal{P}_{image}(iO) \) denotes the prediction derived using image data
    \item \( \alpha \) and \( \beta \) are the learnable parameters that get multiplied with the output of the respective feature extractors before passing through the predictors in the forward pass. This is done to ensure that the parameters become learnable by getting their gradients updated. In the end, the updated \( \alpha \) and \( \beta \) parameters are multiplied with the predictions from the image and meteorological predictors, respectively, to get the final weighted \textit{VWC} as output.
\end{itemize}

This method is designed to allow the model to automatically focus more on the data modality that is performing better, thereby adapting the prediction mechanism based on the intrinsic value of each type of data. By optimizing \( \alpha \) and \( \beta \) within the training process, the model flexibly adapts to the underlying patterns and relationships in the dataset, enhancing both accuracy and robustness in predictions.

To the best of our knowledge, this approach of using dual learnable parameters for dynamic feature weighting has not been explored in previous studies. This approach not only boosts the model’s adaptability and performance but also provides deeper insights into the relative importance of different data sources in predicting soil moisture. Fig.~\ref{fig:lpplot} illustrates the adaptive weighting mechanism, highlighting how \( \alpha \) and \( \beta \) influence the final prediction based on the effectiveness of the data modalities.


%% file: 5-Results.tex
\section{Results \& Discussion}
In this section, we evaluate our proposed three-fold MIS-ME framework with existing baselines leveraging evaluation metrics like Mean Absolute Error (MAE) and Mean Absolute Percentage Error (MAPE).

\subsection{Dataset Splitting Scheme}
We have combined the data from all three stations to form the training, validation, and test sets in the ratio 65:15:20. Apart from evaluating the combined test set, the 20\% test samples from each station have been evaluated separately as well to get an insight into the station-wise performance of the model. The validation set has played a crucial role in the hyperparameter tuning process of the models. 

\subsection{Performance Analysis of Meteo-Only Models}
We assess the performance of several baseline models alongside our MSME method trained on tabular meteorological data only, as depicted in Table~\ref{tab:regression}. SVR performs unfavorably with a MAPE of 16.79\%. In contrast, deep learning models such as the Transformer and 1D-CNN offer more competitive results, with MAPEs of 13.56\% and 14.38\%, respectively. LSTMs and their variants, including CNN-with-LSTM and FTA-LSTM (an LSTM combining both feature and temporal attention mechanisms), also perform robustly with CNN-with-LSTM outperforming all other baselines with a MAPE of 13.39\%. Our MSME architecture, integral to our MIS-ME framework, achieves a MAPE of 15.26\%, demonstrating its effectiveness for further exploration with multi-modal soil moisture estimation strategies. All three MIS-ME approaches in Table~\ref{tab:regression} use the MSME model as feature extractor for the meteo data.

\subsection{Performance Analysis of Image-Only Models}
In evaluating image-only models for soil moisture estimation, we observe varied performance across several architectures as shown in Table~\ref{tab:regression}. MobileNetV2 stands out with the lowest MAPE of 12.3\%, likely due to its efficient design that balances complexity and performance. This makes it particularly effective for processing soil patch images. In comparison, ResNet18 shows decent results with a MAPE of 15.95\%. EfficientNetV2 and InceptionV3 do not perform as well, recording higher MAPEs of 19.35\% and 19.62\%. 
Overall, the performance disparity among these models highlights the importance of choosing the right architecture based on the specific nature and scale of the data involved.
\begin{table}[tbp]
\centering
\caption{Result Comparison of MIS-ME with Baselines}
\label{tab:regression}
\begin{tblr}{
  cells = {c},
  cell{2}{1} = {r=7}{},
  cell{9}{1} = {r=4}{},
  cell{13}{1} = {r=4}{},
  cell{17}{1} = {r=4}{},
  cell{21}{1} = {r=4}{},
  vlines,
  hline{1-2,25} = {-}{},
  hline{9,13,17,21} = {-}{dashed},
}
\textbf{Approach}                     & \textbf{Model type}  & \textbf{MAE}   & \textbf{MAPE}  \\
Only Meteo Data                       & SVR \cite{cnn-with-lstm}                 & 0.044          & 16.79          \\
                                      & MSME                 & 0.04           & 15.26          \\
                                      & 1D-CNN \cite{cnn-with-lstm}             & 0.042          & 14.38          \\
                                      & LSTM \cite{lstm}                & 0.043          & 15.31          \\
                                      & CNN-with-LSTM \cite{cnn-with-lstm}        & 0.038          & 13.39          \\
                                      & FTA-LSTM \cite{fta-lstm}             & 0.044          & 14.86          \\
                                      & Transformer \cite{transformer}         & 0.038          & 13.56          \\
Only Image Data                       & ResNet18 \cite{resnet18}            & 0.043          & 15.95          \\
                                      & MoblieNetV2 \cite{mobilenetv2}          & 0.032          & 12.3           \\
                                      & EfficientNetV2 \cite{efficientnetv2}       & 0.046          & 19.35          \\
                                      & InceptionV3 \cite{inceptionv3}          & 0.051          & 19.62          \\
{MIS-ME \\with \\Multimodal Concat}  & ResNet18             & 0.033          & 12.5           \\
                                      & \textbf{MoblieNetV2} & \textbf{0.029} & \textbf{10.99} \\
                                      & EfficientNetV2       & 0.038          & 13.52          \\
                                      & InceptionV3          & 0.037          & 13.74          \\
{MIS-ME \\with \\Hybrid Loss}         & ResNet18             & 0.033          & 12.45          \\
                                      & \textbf{MoblieNetV2} & \textbf{0.026} & \textbf{10.14} \\
                                      & EfficientNetV2       & 0.035          & 13.65          \\
                                      & InceptionV3          & 0.035          & 14.04          \\
{MIS-ME \\with \\Learnable Parameter} & ResNet18             & 0.034          & 12.85          \\
                                      & \textbf{MoblieNetV2} & \textbf{0.029} & \textbf{11.16} \\
                                      & EfficientNetV2       & 0.036          & 13.44          \\
                                      & InceptionV3          & 0.035          & 13.91          
\end{tblr}
\end{table}

\subsection{Performance Analysis of MIS-ME}
\subsubsection{MIS-ME with Multi-modal Concat}
In examining the effectiveness of the Multi-modal Concat approach within our MIS-ME framework, significant performance improvements are observed when compared to the unimodal approaches. As outlined in Table~\ref{tab:regression}, the Multi-modal Concat strategy using MobileNetV2 achieves a notable MAPE of 10.99\% and outperforms all image-only and meteo-only models. 
This improvement underscores the impact of integrating both meteorological and image data, where the concatenated approach leverages information from both modalities, thereby resulting in more accurate and robust soil moisture predictions.

\begin{figure*}[htbp]
   \centerline{\includegraphics[scale=0.50]{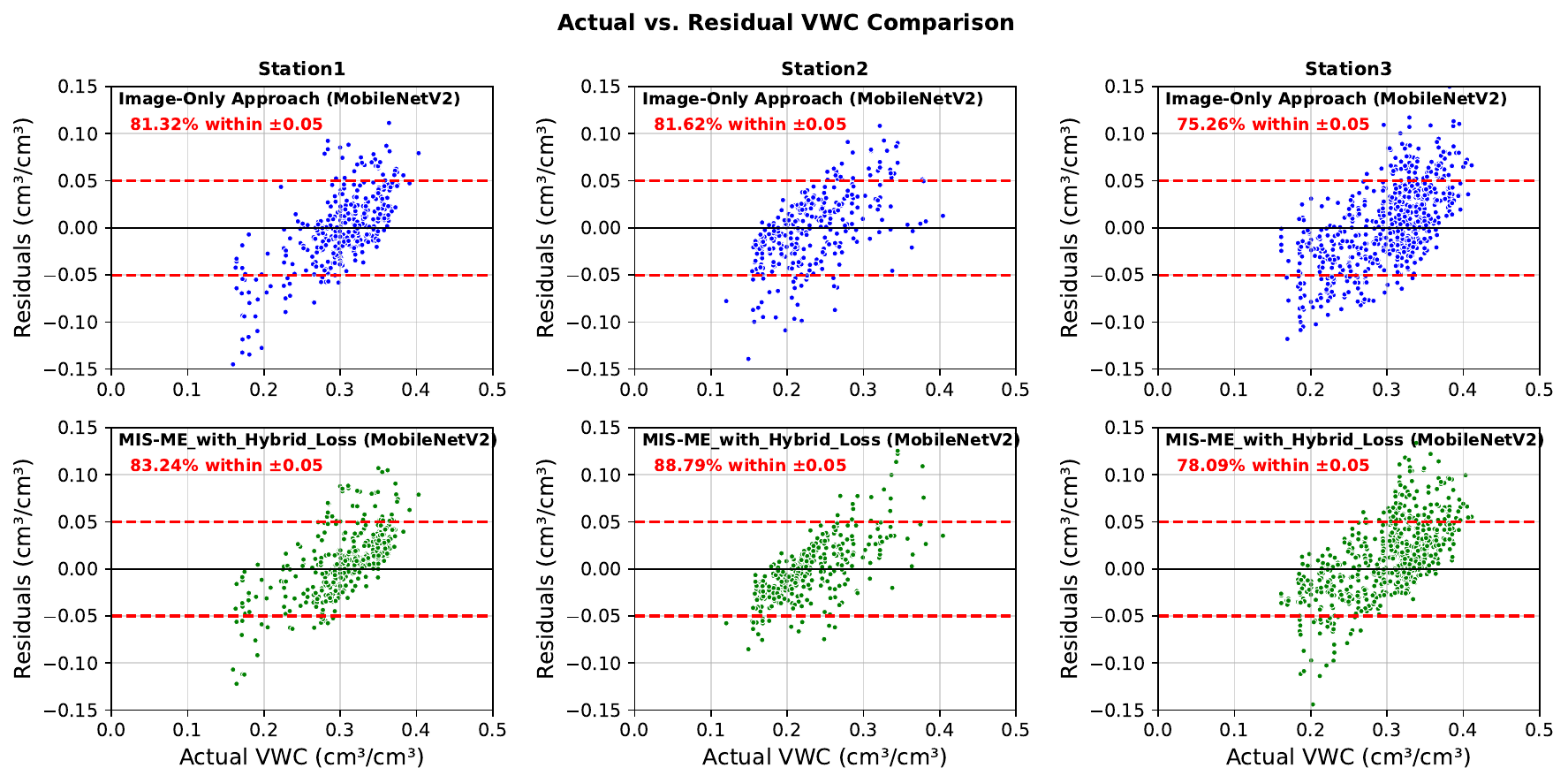}}
   \caption{Residual analysis of the Image-only approach and MIS-ME with Hybrid Loss across three stations for MobileNetV2. Higher percentage of residuals lie within the optimal range for MIS-ME than the regular image-only approach.}
   \label{fig:residualplot}
\end{figure*}

\subsubsection{MIS-ME with Hybrid Loss}

The hybrid loss approach within our MIS-ME framework emerges as the most effective model, achieving the lowest MAPE of 10.14\% with the MobileNetV2 architecture, as shown in Table~\ref{tab:regression}. This method finely balances the loss contributions from concatenated data, meteorological data, and image data. It not only outperforms the unimodal approaches but also improves over the Multi-modal Concat strategy. The weighing coefficients for concatenated loss (\(\delta\)), meteorological loss (\(\gamma\)), and image loss(\(\lambda\)) were all set to 1 to get the best results. Details on the varying of these weighting coefficients are discussed in Section \ref{hybrid-loss-ablation}. This method highlights the significance of customized loss management in multi-modal learning frameworks, making the Hybrid Loss model the top pick of our three-way MIS-ME framework for soil moisture estimation. 

\subsubsection{MIS-ME with Learnable Parameter}
The Learnable Parameter approach within our MIS-ME framework allows for dynamic weighting of meteorological and image-derived features, adapting the model's reliance on each data type based on their predictive value. Table~\ref{tab:regression} shows that this approach achieves a MAPE of 11.16\% with the MobileNetV2 architecture, which is competitive with the other multimodal approaches and significantly outperforms the unimodal models. 
Notably, after training to convergence, the learned weights usually stabilize at \(\alpha = 0.65\) and \(\beta = 0.04\), signifying the greater impact of meteorological data in our dataset. This differential weighting confirms the nuanced integration capability of the learnable parameter approach, effectively harnessing the strengths of both modalities to enhance prediction accuracy.

\vspace{4pt}
\noindent
Overall, the three-way MIS-ME framework outperforms all unimodal approaches convincingly, with the MobileNetV2 architecture coming out on top across all three approaches. Moreover, we see a significant improvement of more than 5\% MAPE for EfficientV2 and InceptionV3 across all three approaches of the MIS-ME framework when compared to training these architectures solely using image data.

\subsection{Station-wise Analysis of MIS-ME}

\subsubsection{Residual Analysis}
Fig.~\ref{fig:residualplot} presents the residual analysis of the Image-only approach and MIS-ME with Hybrid Loss across three stations for MobileNetV2. The analysis shows that most residuals lie within the ideal [-0.05, 0.05] range. Notably, the MIS-ME-trained model demonstrates a higher concentration of residuals within this range than the Image-only approach. For instance, in the case of \emph{Station1}, 83.24\% of MIS-ME's residuals fall within this interval, in comparison to 81.32\% for the image-only approach, reflecting a 2\% enhancement. Similarly, for \emph{Station2}, around 89\% of MIS-ME's residuals are within the range versus 82\% for MobileNetV2, a 7\% improvement. For \emph{Station3}, approximately 78\% of MIS-ME's residuals are within the range, as opposed to 75\% for MobileNetV2, marking a 3\% improvement. These results indicate a tighter clustering of residuals near zero and a better model fit for the MIS-ME-trained model compared to the Image-only MobielnetV2 model.

\begin{figure}[!bp]
\centerline{\includegraphics[scale=0.30]{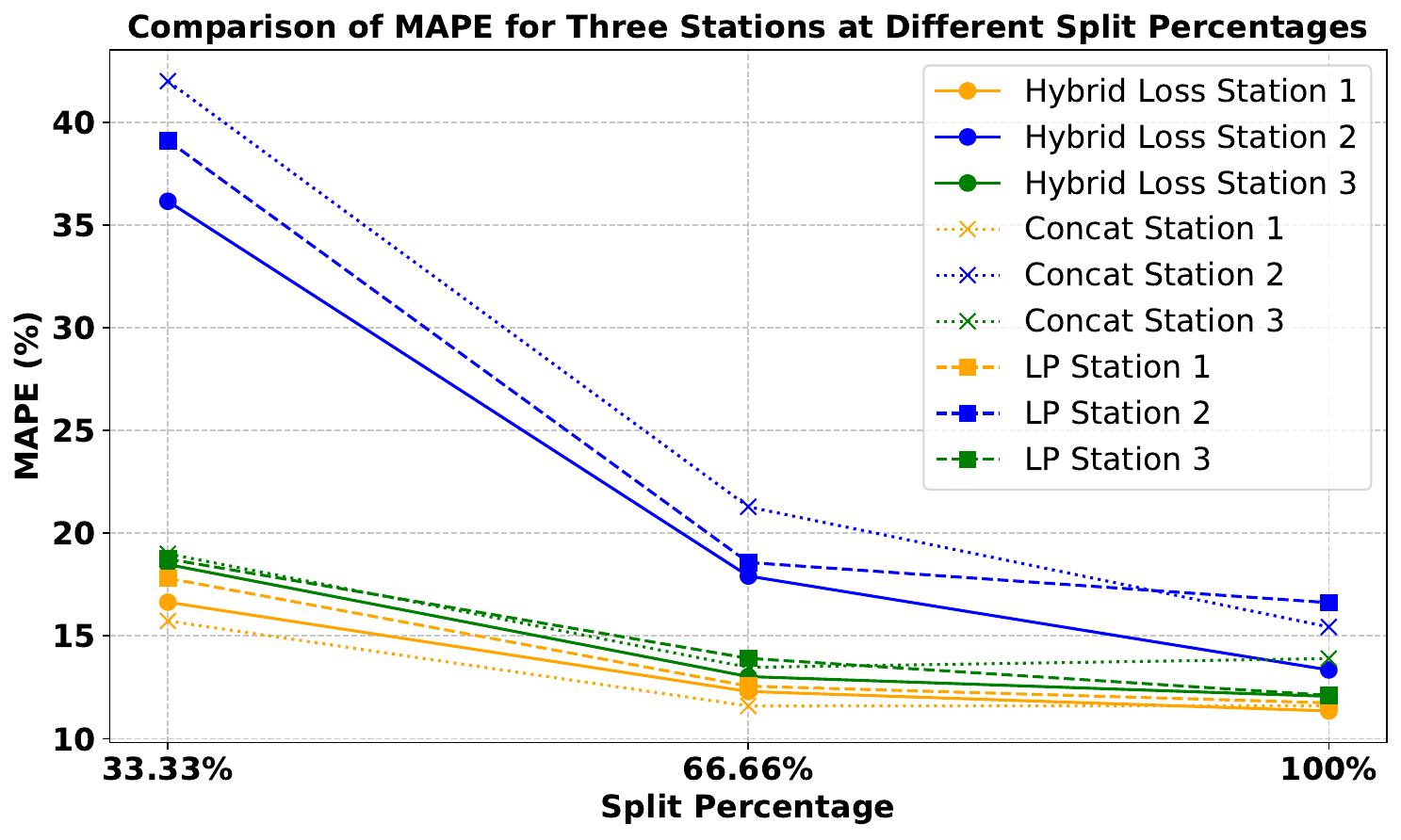}}
\caption{Performance of all three approaches of the MIS-ME framework improves by increasing percentage (\%) of the target station's data in the training sample across all stations. This showcases the requirement of geographic-specific features for a generalized model.}
\label{fig:varying_station}
\end{figure}

\subsubsection{Varying the Target Station Data}
Fig.~\ref{fig:varying_station} illustrates the impact of incrementally introducing data from the target station to the training sample for all three approaches of our MIS-ME framework using MobileNetV2. This approach involves initially training the model using data exclusively from the non-target stations. For example, when targeting \emph{Station1}, the model is trained solely with data from \emph{Station2} and \emph{Station3}, with no data from \emph{Station1} included. This process is then progressively adjusted by adding 33.33\%, 66.66\%, and finally 100\% of \emph{Station1}'s training data, each time evaluating the model's performance on \emph{Station1}'s test set. The results reveal that both \emph{Station1} and \emph{Station3} exhibit similar trends, with MAPE significantly reducing from around 18-16\% to a stable range of 11-12\% for all three approaches as more data from the target station is incorporated. In contrast, \emph{Station2} shows a distinct pattern, starting with a high MAPE of around 36-40\% when its data is initially absent during training, suggesting significantly different soil characteristics compared to the other stations. However, as \emph{Station2} data is gradually introduced, the model's accuracy improves remarkably, stabilizing at a MAPE of 13-16\%. This experiment shows that similar patterns are required for optimal performance when predicting soil moisture in new locations.

\begin{figure}[tbp]
\centering
\centerline{\includegraphics[scale=0.30]{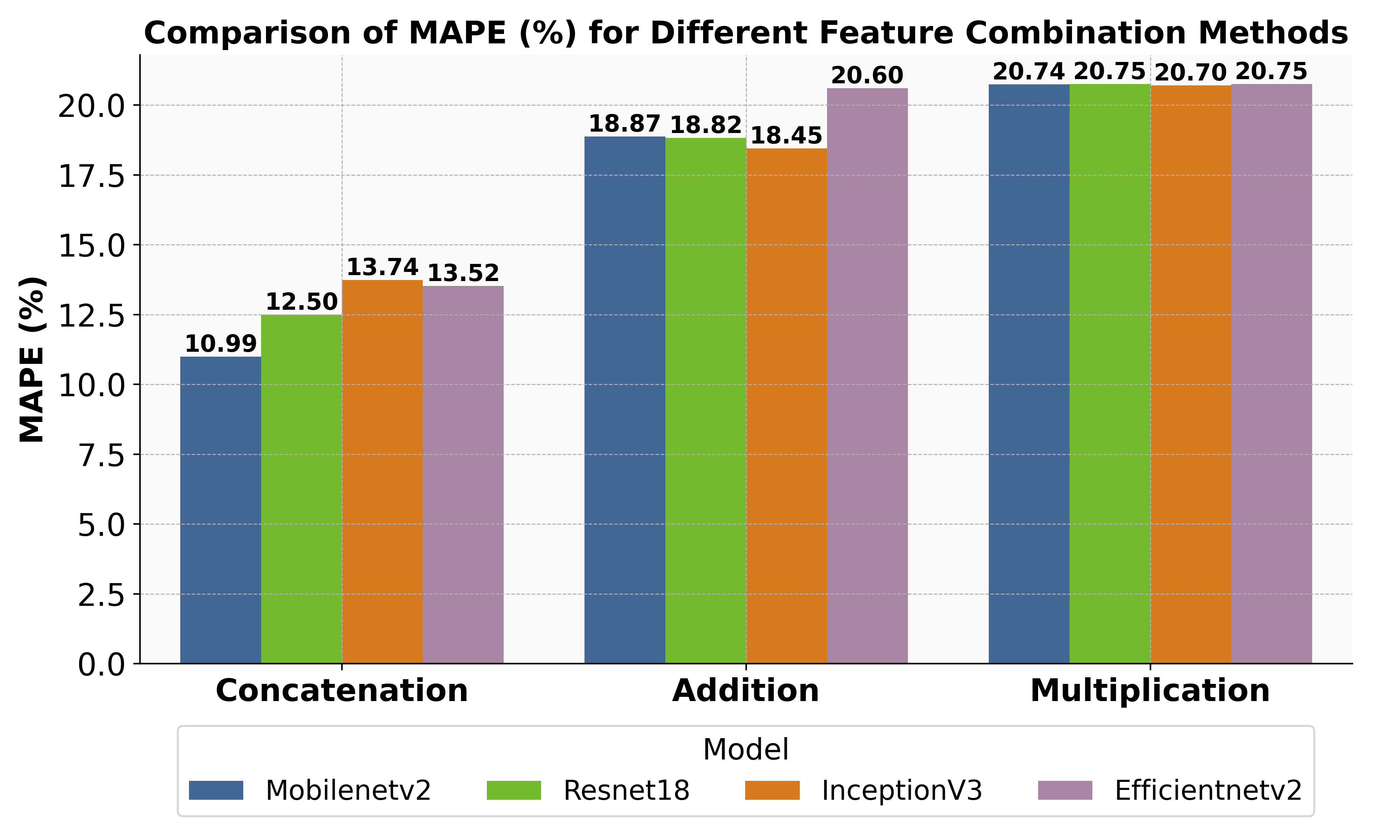}}
\caption{Comparison of MAPE (\%) for Different Feature Combination Methods}
\label{fig:fcmethods}
\end{figure}

\subsection{Ablation Study}

\subsubsection{Feature Combination Methods for MIS-ME with Multimodal Concat}
We explore the performance of different feature combination methods within the MIS-ME framework—specifically concatenation, addition, and multiplication. To prevent dimension mismatch for addition and multiplication, the image feature dimensions were downscaled to match those of meteorological features by passing them through two linear layers, each followed by batch normalization and ReLU activation, with a dropout layer incorporated after the first activation to reduce overfitting. As illustrated in Fig.\ref{fig:fcmethods}, direct concatenation demonstrates the best performance for all architectures. In contrast, addition and multiplication yield poorer results, with multiplication further lagging behind. These results show the effectiveness of concatenation in utilizing complementary information from both modalities, thus enhancing the model's predictive accuracy in soil moisture estimation. For this reason, we have gone with concatenation in our MIS-ME framework.

\subsubsection{Varying Weighting Coefficients \(\delta\), \(\gamma\), and \(\lambda\) in MIS-ME with Hybrid Loss}

\begin{figure}[tp]
\centerline{\includegraphics[width=1.0\linewidth]{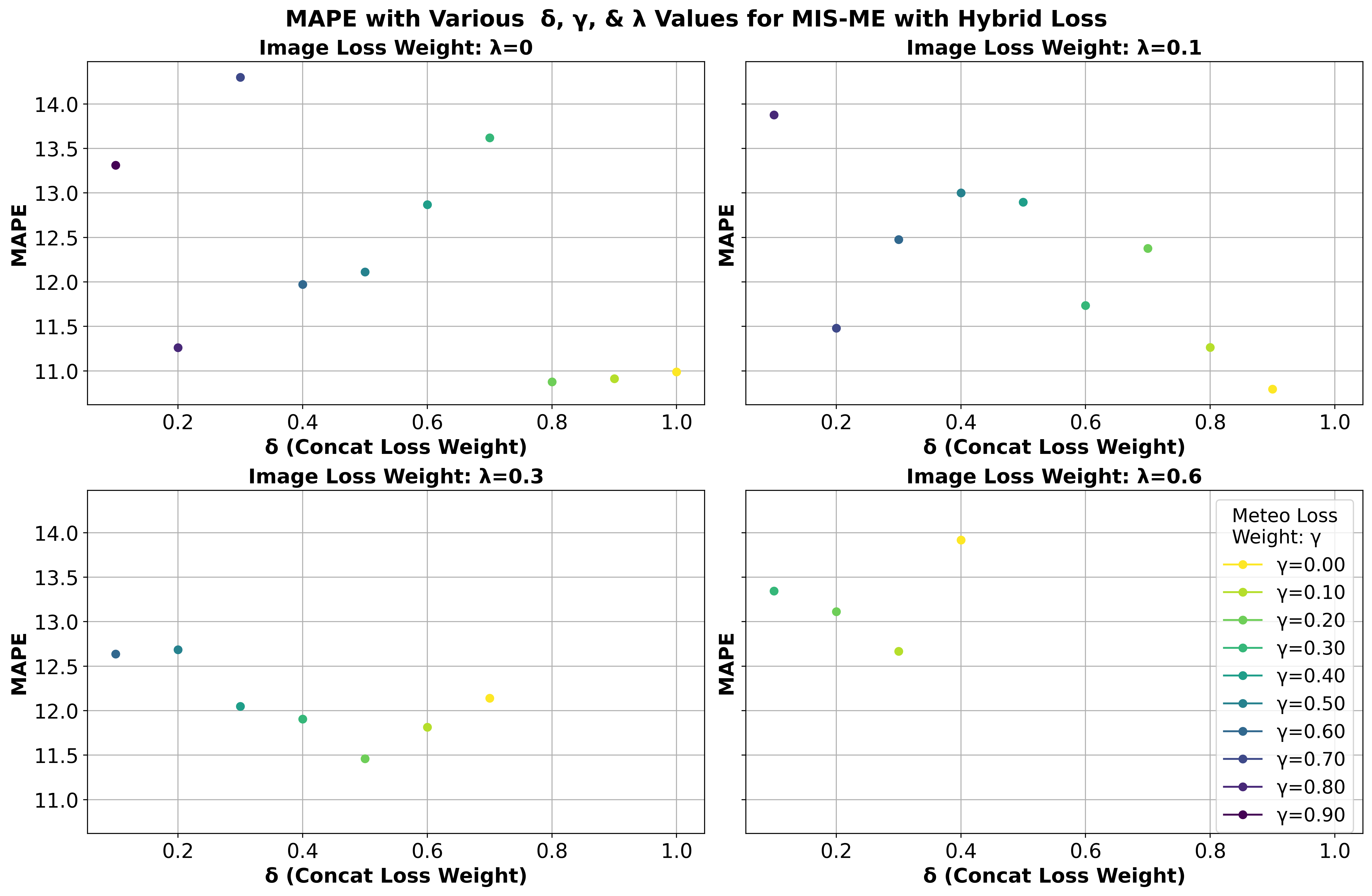}}
\caption{Effect of Varying Loss Coefficients $\delta$, $\gamma$, and $\lambda$.}
\label{fig:hybridloss_ablation}
\end{figure}

In Fig.~\ref{fig:hybridloss_ablation}, we explore the impact of various combinations of \(\delta\), \(\gamma\), and \(\lambda\). We observe that, setting all the coefficients to 1 leads to the best result which is reported in Table~\ref{tab:regression}. Additionally, high weightage to concatenated loss (\(\delta\)) coupled with low weightage to meteo loss (\(\gamma\)) or image loss (\(\lambda\)) also leads to competitive results. We achieve an MAPE of 10.79\% with \(\delta=0.9\), \(\gamma=0\), and \(\lambda= 0.1\). Other effective combinations include \(\delta=0.9\), \(\gamma = 0.1\), and \(\lambda= 0\) or \(\delta = 0.8\), \(\gamma = 0.2\), and \(\lambda= 0\).

\begin{figure}[bp]
\centerline{\includegraphics[scale=0.40]{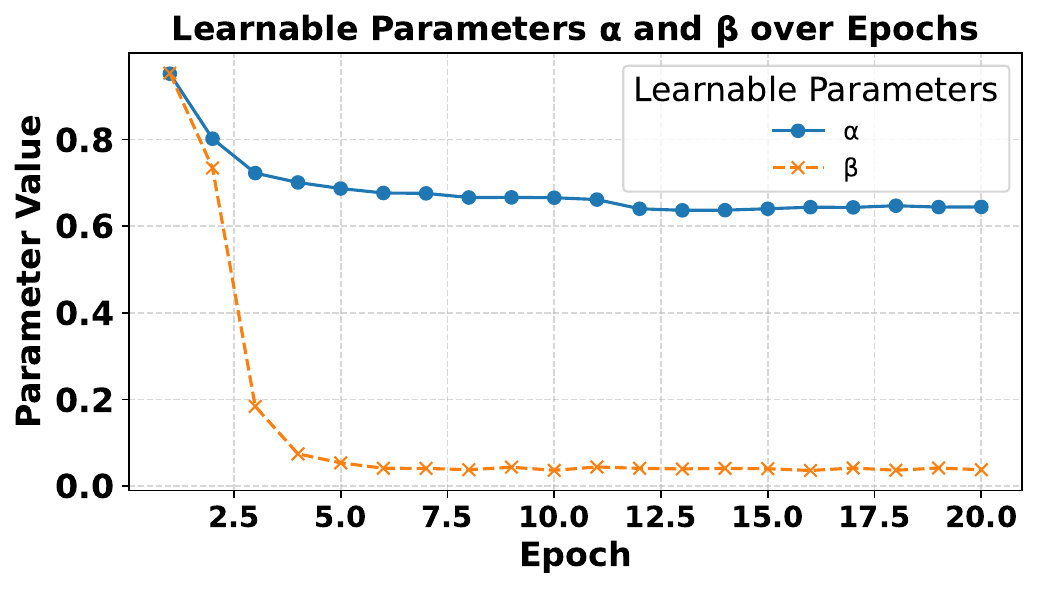}}
\caption{Observing \(\alpha\) and \(\beta\) of MIS-ME-with-Learnable-Parameters over Epochs}
\label{fig:ablation2}
\end{figure}

\subsubsection{Impact of Learnable Parameters on Prediction}
\label{hybrid-loss-ablation}

\paragraph{Analysis of Learned Weights in Enhancing Model Flexibility}
Fig.\ref{fig:ablation2} illustrates the evolution of the learnable parameters \(\alpha\) and \(\beta\) over the training epochs for the best-performing MobileNetV2 model. \(\alpha\) and \(\beta\) dynamically adjust the contributions of the meteorological and image data, respectively, to obtain the final soil moisture prediction. Initially, both \(\alpha\) and \(\beta\) start at 1.0 but gradually decrease as the model learns the optimal weights. By the end of training, \(\alpha\) settles at approximately 0.65, and \(\beta\) at around 0.04. This indicates a higher reliance on meteorological data for accurate predictions in our dataset. Although this plot is specific to MobileNetV2, similar trends are observed across other architectures trained on our dataset. The dynamic adjustment of \(\alpha\) and \(\beta\) enhances model flexibility by prioritizing the modality more relevant to soil moisture labels, thereby improving overall performance in soil moisture estimation.

\paragraph{Single vs. Dual Learnable Parameters}
In our MIS-ME framework, we utilized two independent learnable parameters, \(\alpha\) and \(\beta\), for the meteorological and image modalities, respectively. To further explore the effectiveness of this approach, we conducted an ablation study using a single learnable parameter, \(\alpha\), for the meteorological modality and \(1-\alpha\) for the image modality, forcing the model to learn a complementary relationship between them. However, results indicated that using such complementary parameters led to slightly higher MAPEs than independent learnable parameters. This suggests that allowing the model to independently adjust the weight of each modality offers better flexibility and performance for soil moisture estimation in our dataset.

%% file: 6-Conclusion.tex
\section{Conclusion}
In this work, we explore machine-learning approaches with real-time images and tabular meteorological data collected from three ground stations for the soil moisture regression task. We highlight that this dataset is collected from the wild and resembles images taken from mobile phones. This makes them useful for evaluating real-time soil moisture estimation with commonly available images.
We demonstrate that integrating image features with meteorological data using our proposed three-way MIS-ME framework significantly improves the performance of soil moisture regression compared to traditional approaches, which rely only on meteorological data. Notably, we experimented with feature combination techniques, hybrid loss functions, and learnable parameters. In the future, we aim to experiment with other methods, such as employing transfer-learning techniques like knowledge distillation. We conclude that including features of soil patches can positively impact the estimation of ground soil moisture, opening a new arena of research in the computational agriculture domain.


